\documentclass[letterpaper]{article}
\usepackage{aaai}
\usepackage{times}
\usepackage{helvet}
\usepackage{courier}
\usepackage{graphicx} 
\usepackage{subfigure} 

\usepackage{amsfonts}

\usepackage{url}
\usepackage{booktabs}
\usepackage{rotating}
\usepackage{multirow}
\usepackage{colortbl}
\usepackage{amsmath}

\newtheorem{thm}{Theorem}
\newtheorem{example}{Example}
\newtheorem{definition}[thm]{Definition}

\newcommand{\argmax}{\operatornamewithlimits{arg\ max}}
\newcommand{\argmin}{\operatornamewithlimits{arg\ min}}

\begin{document}
%
\title{RockIt: Exploiting Parallelism and Symmetry for MAP Inference in \\Statistical Relational Models}
\author{
Jan Noessner\\
Data and Web Science Research Group\\
University of Mannheim\\
Mannheim, 68131, Germany\\
\And
Mathias Niepert\\
Computer Science \& Engineering\\
University of Washington\\
Seattle, WA 98195-2350, USA\\
\And Heiner Stuckenschmidt\\
Data and Web Science Research Group\\
University of Mannheim\\
Mannheim, 68131, Germany\\
}
\maketitle

\begin{abstract} 
\begin{quote}
\textsc{RockIt} is a maximum a-posteriori (MAP) query engine for statistical relational models.
MAP inference in graphical models is an optimization problem which can be compiled to integer linear programs (ILPs). 
We describe several advances in translating MAP queries to ILP instances  and present the novel meta-algorithm cutting plane aggregation (CPA).
CPA exploits local context-specific symmetries and bundles up sets of  linear constraints. The resulting counting constraints lead to more compact ILPs and make the symmetry of the ground model more explicit to state-of-the-art ILP solvers. Moreover, \textsc{RockIt} parallelizes most parts of the MAP inference pipeline taking advantage of ubiquitous shared-memory multi-core architectures.

We report on extensive experiments with Markov logic network (MLN) benchmarks showing that \textsc{RockIt} outperforms the state-of-the-art systems \textsc{Alchemy}, \textsc{Markov TheBeast}, and \textsc{Tuffy} both in terms of efficiency and quality of results.
\end{quote}
\end{abstract} 

\section{Introduction}

Maximum a-posteriori (MAP) queries in statistical relational models ask for a most probable possible world given evidence. In this paper, we will present novel principles and algorithms for solving MAP queries in SRL models. MAP inference (or, alternatively, MPE inference) is an important type of probabilistic inference problem in graphical models which is also used as a subroutine in numerous weight learning algorithms~\cite{lowd:2007}. Being able to answer MAP queries more efficiently often translates to improved learning performance. 

Since Markov logic~\cite{richardson2006markov} is arguably the most widely used statistical relational learning (SRL) formalism, we use Markov logic as representation formalism throughout this paper. However, the proposed approach is also applicable to numerous alternative SRL formalisms.
Due to its expressiveness and declarative nature, numerous real-world problems have been modeled with Markov logic. Especially in the realm of data management applications such as entity resolution~\cite{singla2006entity}, data integration~\cite{niepert2010probabilistic,niepert:2010}, ontology refinement \cite{wu2008automatically}, and information extraction~\cite{poon2007joint,kok2008extracting}, Markov logic facilitates rapid prototyping and achieves competitive empirical results. 

The main contributions of the presented work are as follows. 
First, we present a more compact compilation of MAP problems to integer linear programs (ILPs) where each ground clause is modeled by a single linear constraint.
Secondly, we introduce cutting plane aggregation (CPA). CPA exploits evidence-induced symmetries in the ground model that lead to context-specific exchangeability of the random variables. CPA results not only in more compact ILPs but it also makes the model's symmetries more explicit to symmetry detection heuristics of state of the art ILP solvers. 
Thirdly, we parallelize most parts of the MAP query pipeline so as to leverage multi-core architectures.
Finally, we have designed and implemented the presented theory resulting in a novel and robust MLN engine \textsc{RockIt} that integrates cutting plane aggregation with cutting plane inference. 
Numerous experiments on established benchmarks show that CPA leads to significantly reduced run times. \textsc{RockIt} outperforms the state-of-the-art systems \textsc{Alchemy}, \textsc{Markov TheBeast}, and \textsc{Tuffy} both in terms of running time and quality of results on each of the MLN benchmarks.

\section{Related Work}

MaxWalkSAT (MWS), a random walk algorithm for solving weighted SAT problems~\cite{kautz1997general}, is the standard inference algorithm for MAP queries in the Markov logic engine \textsc{Alchemy}~\cite{Alchemy}. The system \textsc{Tuffy}~\cite{niu2011tuffy} employs relational database management systems to ground Markov logic networks more efficiently. \textsc{Tuffy} also runs MWS on the ground model which it initially attempts to partition into disconnected components.

MAP queries in SRL models can be formulated as integer linear programs (ILPs). In this context, cutting plane inference (CPI) solving multiple smaller ILPs in several iterations has shown remarkable performance~\cite{riedel2008improving}. In each CPI iteration, only the ground formulas violated by the current intermediate solution are added to the ILP formulation until no violated ground formulas remain. Since CPI \emph{ignores} ground formulas satisfied by the evidence, it can be seen as a generalization of pre-processing approaches that count the formulas satisfied by the evidence~\cite{shavlik:2009}. In the context of max-margin weight learning for MLNs~\cite{huynh:2009} the MAP query was formulated as a linear relaxation of an ILP and a rounding procedure was applied to extract an approximate MAP state. \textsc{RockIt}'s ILP formulation requires less constraints for ground clauses with negative weights and it combines CPI with cutting plane aggregation.

There is a large class of symmetry-aware algorithms for SRL models. Examples of such \emph{lifted inference} algorithms include first-order variable elimination (\textsc{FOVE})~\cite{poole:2003} and some of its extensions~\cite{milch:2008,kisynski2009lifted} making use of counting and aggregation parfactors. \textsc{FOVE} has also been adapted to solve MAP problems (\textsc{FOVE-P})~\cite{braz:2006}. \cite{apsel:2012} introduced an approach for MAP inference that takes advantage of uniform assignments which are groups of random variables that have identical assignments in some MAP solution. Automorphism groups of graphical models were used to lift variational approximations of MAP inference~\cite{bui:2012}. We attempted to compute automorphism groups as an alternative method for aggregating constraints but experiments showed that calling a graph automorphism algorithm in each CPI iteration dominated the overall solving time. 
\cite{mladenov:2012} computed approximate  solutions to linear programs by reducing the LP problem to a pairwise MRF over  Gaussians and applying lifted Gaussian belief propagation. Similar to the approach of~\cite{bui:2012} lifted linear programming can be used to approximate LP relaxations \cite{asano2006improved} of the MAP ILP. Contrary to previous work, \textsc{RockIt} uses a more compact ILP formulation with a one-to-one correspondence between ground clauses and linear constraints, tightly integrates CPI and CPA, and  \emph{estimates} the optimal aggregation scheme avoiding a costly exact computation in each CPI iteration. Moreover, contrary to lifted inference approaches operating solely on the first-order level, \textsc{RockIt} exploits  evidence-induced local symmetries on the ground level.

There are several lifted \emph{marginal} inference approaches such as lifted message passing~\cite{Singla:2008,kersting:2009}, variants of lifted knowledge compilation and theorem proving~\cite{broeck:2011,gogate2011probabilistic}, and lifted MCMC~\cite{niepert:2012,venugopal:2012} approaches.
While there are some generic parallel machine learning architectures such as \textsc{GraphLab}~\cite{low:2010} which could in principle be used for parallel MAP inference, \textsc{RockIt} is the first system that parallelizes MAP inference in SRL models combining CPI and CPA.

\subsection{Markov Logic}


Markov logic is a first-order template language combining first-order logic with log-linear  graphical models.
We first review function-free first-order logic~\cite{genesereth1987logical}. 
Here, a term is either a constant or a variable. An atom $p(t_1, ... ,t_n)$ consists of a predicate $p/n$ of arity $n$ followed by $n$ terms $t_i$. A literal $\ell$ is an atom $a$ or its negation $\neg a$.  A clause is a disjunction $\ell_1 \vee ... \vee \ell_k$ of literals. The variables in clauses are always assumed to be universally quantified. The Herbrand base $\mathcal{H}$ is the set of all possible ground (instantiated) atoms. 
Every subset of the Herbrand base is a Herbrand interpretation. 

A Markov logic network $\mathcal{M}$ is a finite set of pairs $(F_i, w_i), 1 \leq i \leq n$, where each $F_i$ is a clause in function-free first-order logic and $w_i\in\mathbb{R}$. Together with a finite set of constants $C = \{c_1,...,c_n\}$ it defines the ground Markov logic network $\mathcal{M_{C}}$ with one binary variable for each  grounding of predicates occurring  in $\mathcal{M}$ and
one feature for each grounding of formulas in $\mathcal{M}$ with feature weight $w_i$.
Hence, a Markov logic network defines a log-linear probability distribution over Herbrand interpretations (possible worlds)
\begin{equation}
P(\mathbf{x}) = \frac{1}{Z} \exp\left(\sum_i w_i n_i(\mathbf{x})\right)
\end{equation}
where $n_i(\mathbf{x})$ is the number of satisfied groundings of clause $F_i$ in possible world $\mathbf{x}$ and $Z$ is a normalization constant.


In order to answer a MAP query given evidence $\mathbf{E}=\mathbf{e}$ one has to solve  the  maximization problem
$$\argmax_{\mathbf{x}} \ P(\mathbf{X} = \mathbf{x}\ |\ \mathbf{E} = \mathbf{e})$$
where the maximization is performed over possible worlds (Herbrand interpretations) $\mathbf{x}$ compatible with the evidence.

\section{Cutting Plane Aggregation}
\label{sec:cpa}

Each MAP query corresponds to an optimization problem with linear constraints and a linear objective function and, hence, we can formulate the problem as an instance of integer linear programming. The novel cutting plane aggregation approach is tightly integrated with cutting plane inference (CPI) a meta-algorithm operating between the grounding algorithm and the ILP solver~\cite{riedel2008improving}. Instead of immediately adding one constraint for each ground formula to the ILP formulation, the ILP is initially formulated so as to enforce the given evidence to hold in any solution. Based on the  solution of this more compact ILP one determines the violated constraints, adds these to the ILP, and resolves. This process is repeated until no constraints are violated by an intermediate solution. 

We begin by introducing a novel ILP formulation of MAP queries for Markov logic networks. In contrast to existing approaches~\cite{riedel2008improving,huynh:2009}, the formulation requires only one linear constraint per ground clause irrespective of the ground clause being weighted or unweighted.
Moreover, we introduce the notion of \emph{context-specific exchangeability} and describe the novel \emph{cutting plane aggregation} (CPA) algorithm that exploits this type of local symmetry. Contrary to most symmetry-aware and lifted inference algorithms that assume no or only a limited amount of evidence, the presented approach specifically exploits model symmetries induced by the given evidence. 

\subsection{General ILP Formulation}

In order to transform the MAP problem to an ILP we have to first ground, that is, instantiate, the first-order theory specified by the Markov logic network. Since we are employing cutting plane inference, \textsc{RockIt} runs in each iteration several join queries in a relational database system to retrieve the ground clauses violated by the current solution. Hence, in each iteration of the algorithm, \textsc{RockIt} maintains a set of ground clauses $\mathcal{G}$ that have to be translated to an ILP instance. 

Given such a set of ground clauses $\mathcal{G}$, we associate one binary ILP variable $x_{\ell}$ with each ground atom $\ell$ occurring in some  $g\in\mathcal{G}$. For the sake of simplicity, we will often denote ground atoms and ILP variables with identical names. For a ground clause $g \in \mathcal{G}$ let $L^+(g)$ be the set of ground atoms occurring unnegated in $g$ and $L^-(g)$ be the set of ground atoms occurring negated in $g$.
Now, we encode the given evidence by introducing linear constraints of the form $x_{\ell} \leq 0$ or  $x_{\ell} \geq 1$ depending on whether the evidence sets the corresponding ground atom $\ell$ to false or true.
For every ground clause $g \in \mathcal{G}$ with weight $w > 0, \ w \in \mathbb{R},$ we add a novel binary variable $z_g$ and the following constraint to the ILP:
$$\displaystyle\sum_{\ell \in {L}^{+}(g)} x_{\ell} + \displaystyle\sum_{\ell \in {L}^{-}(g)} (1-x_{\ell}) \geq z_g.$$
Please note that if any of the ground atoms $\ell$ in the ground clause is set to false (true) by the given evidence, we do not include it in the linear constraint. 

For every $g$ with weight $w_g <0, \ w \in \mathbb{R},$ we add a novel binary variable $z_g$ and the following constraint to the ILP:
$$ \displaystyle\sum_{\ell \in {L}^{+}(g)} x_{\ell} + \displaystyle\sum_{\ell \in {L}^{-}(g)} (1-x_{\ell}) \leq (|{L}^{+}(g)|+|{L}^{-}(g)|)  z_g.$$

For every $g$ with weight $w_g = \infty$, that is, a hard clause, we add the following linear constraint to the ILP:
$$\displaystyle\sum_{\ell \in L^{+}(g)} x_{\ell} + \displaystyle\sum_{\ell \in L^{-}(g)} (1-x_{\ell})	\geq 1$$

If a ground clause has zero weight we do not have to add the corresponding constraint.

Finally, the objective of the ILP is:
$$\max \displaystyle\sum_{g \in \mathcal{G}} w_g z_g,$$
where we sum over weighted ground clauses only, $w_g$ is the weight of $g$, and $z_g \in \{0,1\}$ is the binary variable previously associated with ground clause $g$. We compute a MAP state by solving the ILP whose solution corresponds one-to-one to a MAP state $\mathbf{x}$ where $x_i = \texttt{true}$ if the corresponding ILP variable is $1$ and $x_i = \texttt{false}$ otherwise.  

For example, Table~\ref{tab:transformation} depicts three clauses with $w > 0$, $w < 0$, and $w = \infty$, and the respective ILP formulations.

\begin{table}[t!]
\centering
\small
\begin{tabular}{ll}
weight & ground clause \\
 & \\
$1.1$ & $x_1 \vee \neg x_2 \vee x_3$  \\ 
$-0.5$ & $\neg x_1 \vee x_2$ \\ 
$\infty$ &  $\neg x_1 \vee x_2$  \\ 
\end{tabular}
$\ \leadsto \ $
\begin{tabular}{l}
\textbf{max} $1.1 z_1 -0.5 z_2$ \\
\textbf{subject to} \\ 
$x_1 + (1-x_2) + x_3 \geq z_1$ \\
$(1-x_1) + x_2 \leq 2 \cdot z_2$ \\
$(1-x_1) + x_2 \geq 1$\\
\end{tabular}
\caption{\label{tab:transformation}An example of the ILP formulation.}
\end{table}

\subsection{Constraint Aggregation}
\label{sec:aggregation}

In this section we  optimize the compilation of \emph{sets} of weighted ground clauses to \emph{sets} of linear constraints. More concretely, we introduce a novel approach that aggregates sets of ground clauses so as to make the resulting ILP have (a) fewer variables (b) fewer constraints and (c) its context-specific symmetries more exposed to the ILP solver's symmetry detection heuristics.

We first demonstrate that evidence often introduces symmetries in the resulting sets of ground clauses and, therefore, at the level of ILP constraints. 
The proposed approach aggregates ground clauses, resulting in smaller constraint matrices and aiding symmetry detection algorithms of the ILP solvers. The solvers apply heuristics to test whether the ILP's constraint matrix exhibits symmetries in form of permutations of its columns and rows. For a comprehensive overview of existing principles and algorithms for detecting and exploiting symmetries in integer linear programs we refer the reader to~\cite{Margot:2010,Margot:2003,ostrowski:2011,bodi2013algorithms}.
We describe cutting plane aggregation in two steps. First, we explain the aggregation of ground formulas and, second, we describe the compilation of aggregated  formulas to ILP constraints.

\begin{definition}
\label{definition-agg}
Let $G \subseteq \mathcal{G}$ be a set of $n$ weighted ground clauses and let $c$ be a ground clause. We say that $G$ can be aggregated with respect to $c$ if (a) all ground clauses in $G$ have the same weight and (b) for every $g_i \in G, 1 \leq i \leq |G|,$ we have that $g_i = \ell_i \vee c$ where $\ell_i$ is a (unnegated or negated) literal for each $i, 1 \leq i \leq |G|$. 
\end{definition}

\begin{example}
Table~\ref{tab:exAggrClause} lists a set of 5 ground clauses. The set of clauses $\{g_1, g_2, g_3\}$ can be aggregated with respect to $\neg y_1 \vee y_2$ since we can write each of these ground clauses as $\ell_i \vee \neg y_1 \vee y_2$ with $\ell_1:=x_1, \ell_2:=x_2$, and $\ell_3:=\neg x_3$.
\end{example}

Before we describe the advantages of determining ground clauses that can be aggregated and the corresponding ILP formulation encoding these sets of clauses, we provide a typical instance of a Markov logic network resulting in a large number of clauses that can be aggregated.  

\begin{example}
\label{example-cpa}
Let us consider the clause $\neg \mathsf{smokes}(x) \vee \mathsf{cancer}(x)$ and let us assume that there are $100$ constants $C_1,...,C_{100}$ for which we have evidence $\mathsf{smokes}(C_i), 1 \leq i \leq 100$. For $1 \leq i \leq 100,$ let $y_i$ be the ILP variable corresponding to the ground atom $\mathsf{cancer}(C_i)$. 
The naive formulation would contain $100$ constraints $y_i \geq z_i$ and the objective of the ILP with respect to these clauses would be $\max 1.5 z_1 + ... + 1.5 z_{100}$. Instead, we can aggregate the ground clauses $\mathsf{cancer}(C_i), 1 \leq i \leq 100,$ for $c = \texttt{false}$ and $\ell_i = \mathsf{cancer}(C_i), 1 \leq i \leq 100$.
\end{example}

\begin{table}[t!]
\centering
\small
\begin{tabular}{c|c|c|c}
$g$ &$\ell_i$ & $c$  & $w$\\ 
\hline
$g_1$ &     $x_1 \vee$ & $\neg y_1 \vee y_2$ & 1.0\\ 
    \hline
$g_2$ &     $x_2 \vee$ & $\neg y_1 \vee y_2$ & 1.0\\ 
    \hline
$g_3$ & $\neg x_3 \vee$ & $\neg y_1 \vee y_2$ & 1.0\\ 
\hline
$g_4$ & $\neg x_4 \vee$ & $\neg y_1 \vee y_3$ & 1.0\\ 
    \hline
$g_5$ &      $x_5 \vee$ & $\neg y_1 $ & 0.5\\ 
\hline
\end{tabular}
$\ \leadsto \ $
\begin{tabular}{c|c|c}
$\ell_i$ & $c$ & $w$\\ 
\hline
     $x_1 \vee$ & 
\multirow{3}{*}{$\neg y_1 \vee y_2$} & \multirow{3}{*}{1.0}\\ 
     $x_2 \vee$ &  &\\ 
$\neg x_3 \vee$ &  &\\ 
\hline
$\neg x_4 \vee$ & $\neg y_1 \vee y_3$ & 1.0\\ 
\hline
     $x_5 \vee$ & $\neg y_1 $ & 0.5\\ 
\hline
\end{tabular}
\caption{\label{tab:exAggrClause} A set of ground clauses that can be aggregated.}
\end{table}

Let $G \subseteq \mathcal{G}$ be a set of ground clauses with weight $w$ and let  $c$ be a ground clause. Moreover, let us assume that $G$ can be aggregated with respect to $c$, that is, that each $g \in G$ can be written as $\ell_i \vee c$. The \emph{aggregated feature} $f^{G}$ for the clauses $G$ with weight $w$  maps each interpretation $I$ to an integer value as follows
$$ f^{G}(I) = \left\lbrace 
\begin{array}{cl} 
|G| & \ \ if\ \ I \models c \\
|\{\ell_i \vee c \in G\ |\ I \models \ell_i\}| & \mbox{ \ \ otherwise }
\end{array}  \right\rbrace.$$

The feature resulting from the aggregation, therefore, counts the number of literals $\ell_i$ that are satisfied if the ground clause $c$ is not satisfied and returns the number of aggregated clauses otherwise. Please note that an encoding of this feature in a factor graph would require space exponential in the number of  ground atoms even though the feature only has a linear number of possible values. The feature, therefore, is highly symmetric -- each assignment to the random variables corresponding to the unnegated (negated) literals that has the same Hamming weight results in the same feature weight contribution. This constitutes a feature-specific local form of finite exchangeability~\cite{finetti:1972,diaconis:1977} of random variables  induced by the evidence. Therefore, we denote this form of finite exchangeability as \emph{context-specific exchangeability}. Please note that the concept is related to counting formulas used in some lifted inference algorithms~\cite{milch:2008}.
While standard models such as factor graphs cannot represent such symmetric features compactly, one can encode these \emph{counting features} directly with a \emph{constant} number of ILP constraints. We now describe this translation in more detail.

\begin{table}[t!]
\centering
\small
\begin{tabular}{c|c|c}
\multicolumn{3}{c}{ } \\
$\ell_i$ & $c$ & $w$\\ 
\hline
$x_1 \vee$ & \multirow{3}{*}{$\neg y_1$} & \multirow{3}{*}{0.5}\\ 
$x_2 \vee$ &  & \\ 
$x_3 \vee$ &  & \\ 
\hline
$\neg x_1 \vee$ & 
\multirow{3}{*}{$y_1 \vee \neg y_2$} & \multirow{3}{*}{-1.5}\\ 
$\ \ x_2 \vee$ &  & \\ 
$\neg x_3 \vee$ &  & \\ 
\hline
\end{tabular}
$\leadsto$
\begin{tabular}{l}
\textbf{max} $0.5 z_1 -1.5 z_2$ \\
\textbf{subject to} \\ 
\hline
 $x_1 + x_2 + x_3 + 3(1-y_1) \geq z_1$ \\
 $z_1 \leq 3$                   \\ 
  \\
\hline
 $(1-x_1) + x_2 + (1-x_3) \leq z_2$ \\
 $3\cdot y_1 \leq z_2$\\
 $3 \cdot (1-y_2) \leq z_2$\\
\hline
\end{tabular}
\caption{\label{tab:exAggr} Illustration of the constraint aggregation formulation. For the sake of simplicity, we denote the ground atoms and ILP variables with identical names.}
\end{table}

As before, for any ground clause $c$, let $L^+(c)$ ($L^-(c)$) be the set of ground atoms occurring unnegated (negated) in $c$.
We first show the formulation for clauses with positive weights. Let $G \subseteq \mathcal{G}$ be a set of $n$ ground clauses with weight $w >0$ that can be aggregated with respect to $c$, that is, for each $g \in G$ we have that $g = x_i \vee c$ or $g = \neg x_i \vee c$ for some ground atom $x_i$ and a fixed clause $c$. We now add the following two linear constraints to the ILP:
\begin{multline}
 \sum_{(x_i \vee c) \in G} x_i + \sum_{(\neg x_i \vee c) \in G} (1-x_i) \ + \\
 \sum_{\ell \in L^+(c)} n x_{\ell}  + \sum_{\ell \in L^-(c)} n(1-x_{\ell})   \geq z_g
\end{multline}
\hspace{10mm} and
\begin{equation}
 z_g \leq n   
\end{equation}
Linear constraint (2) introduces the novel \emph{integer} variable $z_g$ for each aggregation. Whenever a solution satisfies the ground clause $c$ this variable has the value $n$ and otherwise it is equal to the number of literals $\ell_i$ satisfied by the solution. Since constraint (2) alone might lead to values of $z_g$ that are greater than $n$, the linear constraint (3) ensures that the value of $z_g$ is at most $n$. However, linear constraint (3) only needs to be added if clause $c$ is not the constant \texttt{false}.

\begin{figure}[t!]
\centering
\includegraphics[scale=0.4]{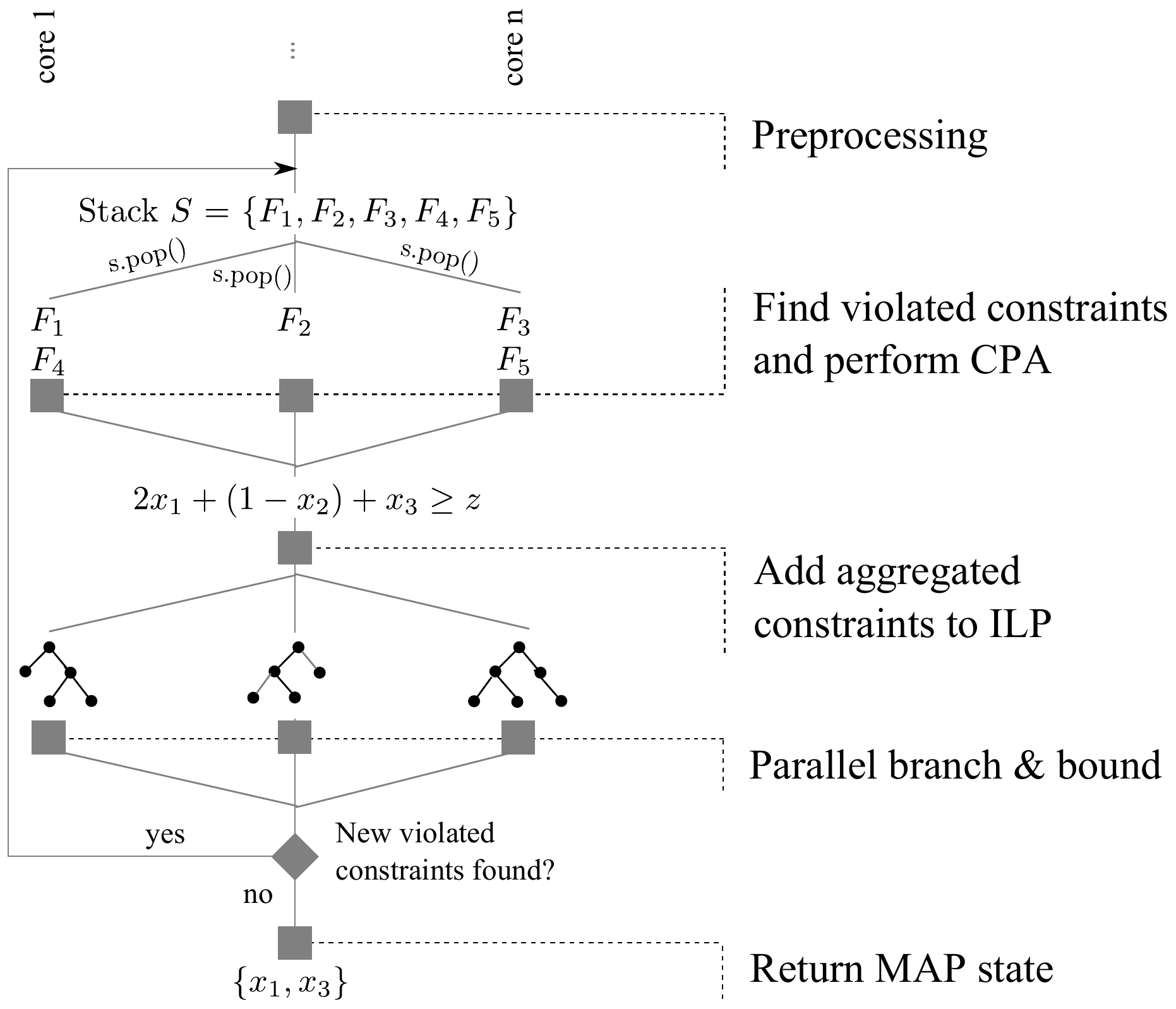}
\caption{\label{fig:parallelization}\textsc{RockIt} parallelizes constraint finding, constraint aggregation, and ILP solving.}
\end{figure}

We describe the aggregation of clauses with negative weight. Let $G \subseteq \mathcal{G}$ be a set of $n$ ground clauses with weight $w < 0$ that can be aggregated with respect to $c$, that is, for each $g \in G$ we have that $g = x_i \vee c$ or $g = \neg x_i \vee c$ for a ground atom $x_i$ and a fixed clause $c$. 
We now add the following linear constraints to the ILP:
\begin{equation}
 \sum_{(x_i \vee c) \in G} x_i + \sum_{(\neg x_i \vee c) \in G} (1-x_i)  \leq z_g
\end{equation}
\hspace{5mm}and
\begin{equation}
n x_{\ell} \leq z_g  \mbox{ for every } \ell \in L^+(c)
\end{equation}
\hspace{5mm}and
\begin{equation}
n (1-x_{\ell}) \leq z_g  \mbox{ for every } \ell \in L^-(c).
\end{equation}

Linear constraint (4) introduces an integer variable $z_g$ that counts the number of ground clauses in $G$ that are satisfied.
For each of the integer variables $z_g$ representing an aggregated set of clauses we add the term $w_g z_g$ to the objective function where $w_g$ is the weight of each of the aggregated clauses. Table~\ref{tab:exAggr} shows a set of aggregated ground clauses and the corresponding ILP formulation. It is not difficult to verify that each solution of the novel formulation corresponds to a MAP state of the MLN it was constructed from.

\begin{example}
In Example~\ref{example-cpa} we aggregate the ground clauses $\mathsf{cancer}(C_i), 1 \leq i \leq 100,$ for $c = \texttt{false}$ and $\ell_i = \mathsf{cancer}(C_i), 1 \leq i \leq 100$. Now, instead of $100$ linear constraints and $100$ summands in the objective function the aggregated formulation only adds the linear constraint $y_1 + ... + y_{100} \leq z_g$ and the term $1.5 z_g$ to the objective. 
\end{example}

We observed that the computation and aggregation of the violated constraints often dominated the ILP solving time. Since, in addition, state-of-the-art ILP solvers such as \textsc{Gurobi} already parallelize their branch and bound based algorithms we developed an additional method for parallelizing the CPI and CPA phases.

\begin{table}
\centering
\small
\begin{tabular}{lrrrrr}
\toprule
&\textbf{ER}&\textbf{IE} & \textbf{LP} &  \textbf{PR} & \textbf{RC} \\
\toprule
 predicates       & 10 		  & 18      &22  		& 7         &4 \\ 
 formulas         & 1,276 	& 1,024   &24  		& 2,461     &17  \\ 
\midrule
 evidence         &12,892 	& 258,079 &1,031 	& 12,999    &99,161 \\ 
 clauses          &390,720  & 340,737 &354,587&40,234,321 &202,215\\ 
\toprule
\multicolumn{6}{c}{Number of ILP constraints}\\
\toprule
w/o CPA         & 357,056 & 4,041      &  31,658  &  2,688,122 & 164,047     \\
w/ \ \ CPA      & 10,782  & 932        &  6,617   &  1,573     & 10,064   \\
\toprule
\end{tabular}
\caption{\label{tab:propConstr} Characteristics of the ML benchmark datasets and the number of constraints of the respective ILP formulations.} 
\end{table}

\section{Parallelism and Implementation}

We will (a) briefly explain the parallelization framework of~\textsc{RockIt}\footnote{http://code.google.com/p/rockit/}, (b) outline the implementation of the aggregation strategy, and (c) provide some details about the third-party components we employed. 

Figure~\ref{fig:parallelization} depicts the computational pipeline of the system. After pre-processing the input MLN and loading it into the relational database system, \textsc{RockIt} performs CPI iterations until no new violated constraints are found. The violated constraints are computed with joins in the relational database system where each table stores the predicate groundings of the intermediate solutions. In each CPI iteration, \textsc{RockIt} performs CPA on the violated constraints. We can parallelize the aggregation steps by processing each first-order formula in a separate thread. To this end, each first-order formula is initially placed on a stack $S$. \textsc{RockIt} creates one thread per available core and, when idle, makes each of the threads (i) pop a first-order formula from the stack $S$, (ii) compute the formula's violated groundings, and (iii) perform CPA on these groundings. The aggregated groundings are compiled into ILP constraints and added to the ILP formulation. When the stack $S$ is empty and all threads idle we solve the current ILP in parallel, obtain a solution, and begin the next CPI iteration.

There are different possible strategies for finding the ground clauses $c$ (see Definition \ref{definition-agg}) that minimize the number of counting constraints per first-order formula. While this problem can be solved optimally with algorithms that detect symmetries in propositional formulas such as \textsc{Saucy}~\cite{darga2008faster} or, alternatively, by reducing it to a frequent itemset mining problem for which several algorithms exist, these algorithms need to be called in each CPI iteration and for each first-order formula and extensive experiments showed that these algorithms dominated ILP solving. Therefore, we implemented a greedy algorithm that only \emph{estimates} the optimal aggregation scheme. The algorithm stores, for each first-order clause, the violated groundings of the form $\ell_{1} \vee \ldots \vee \ell_n$ in a table with $n$ columns where each column represents one literal position of the clause. For each column $k$, \textsc{RockIt} computes the set of \emph{distinct} rows $R_k$ of the table that results from the projection onto the columns $\{1,...,n\}\setminus\{k\}$. Let $d = \argmin_k \{|R_k|\}$. The clause groundings are then aggregated with respect to the rows in $R_{d}$.


\textsc{RockIt} employs \textsc{MySQL}'s in-memory tables for the computation of violated constraints and the aggregation. Most tables are hash indexed to facilitate highly efficient join processing. We use \textsc{Gurobi}\footnote{http://www.gurobi.com/} as \textsc{RockIt}'s internal ILP solver due to its ability to parallelize its branch, bound, and cut algorithm, its remarkable performance on standard ILP benchmarks~\cite{koch2011miplib}, and its symmetry detection heuristics. In addition to the project code, \textsc{RockIt} is made available as a web-service where users can upload MLNs. Furthermore, programmers can integrate the MLN query engine in their own applications via a REST interface.

\section{Experiments}\label{sec:experiments}

With the following experiments we assess whether and to what extent \textsc{RockIt} (a) reduces the number of constraints in the ILP formulation, (b) reduces the overall runtime, and (c) outperforms state-of-the-art MLN systems. We compare \textsc{RockIt} to three MLN systems \textsc{Alchemy}~\cite{Alchemy}, \textsc{Markov TheBeast}~\cite{riedel2008improving}, and  \textsc{Tuffy} (version 3)~\cite{niu2011tuffy}. To ensure a fair comparison, we made \textsc{Markov TheBeast} also use the ILP solver Gurobi.  In addition, we investigate whether and to what extent (d) \textsc{RockIt}'s performance increases with the number of available cores of a shared-memory architecture.
All experiments were performed on a standard PC with $8$ GB RAM and $2$ cores with $2.4$ GHz each unless otherwise stated. 

\begin{figure*}[t!]
\includegraphics[height=\textwidth,angle=270]{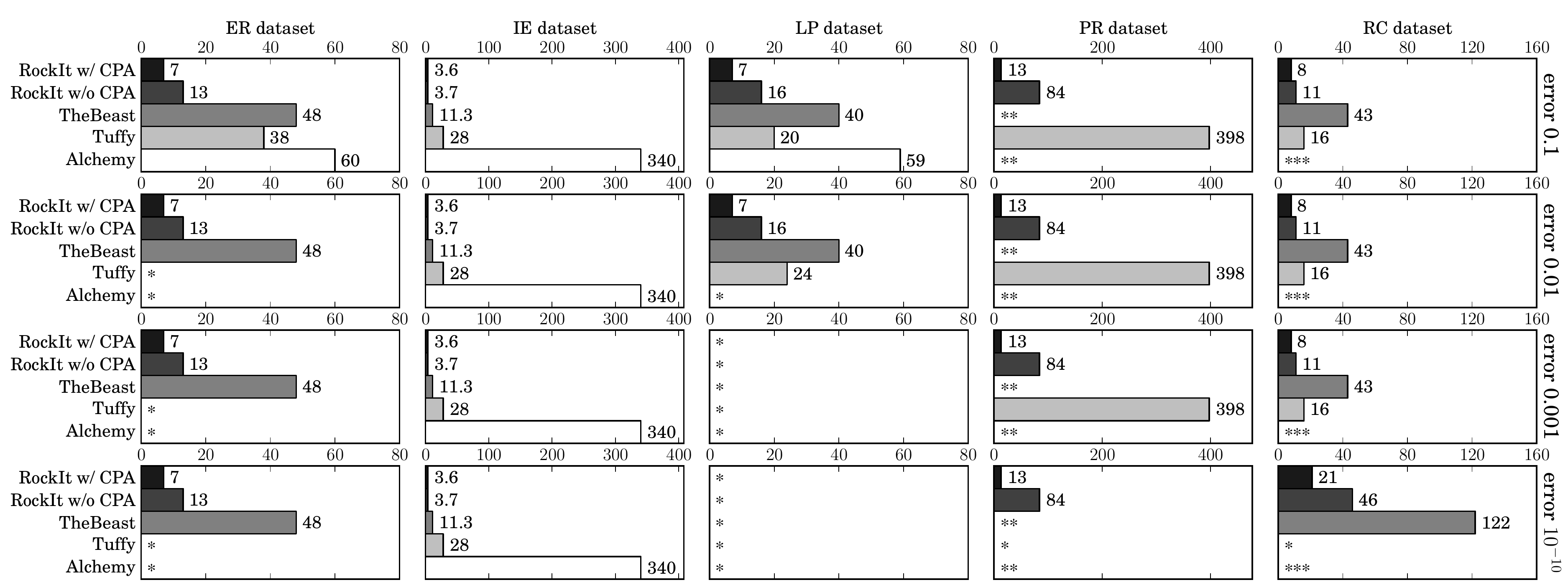}
\caption{\label{fig:runtimesrockIt} Running time in seconds of \textsc{RockIt},  \textsc{Alchemy}, \textsc{Markov TheBeast}, and \textsc{Tuffy} for different gaps (bounds on the relative error) and with two cores. * gap not reached within $1$ hour, ** out of memory, *** did not terminate within 1 hour.}
\end{figure*}

\begin{figure}[t!]
\begin{center}
\includegraphics[width=\linewidth]{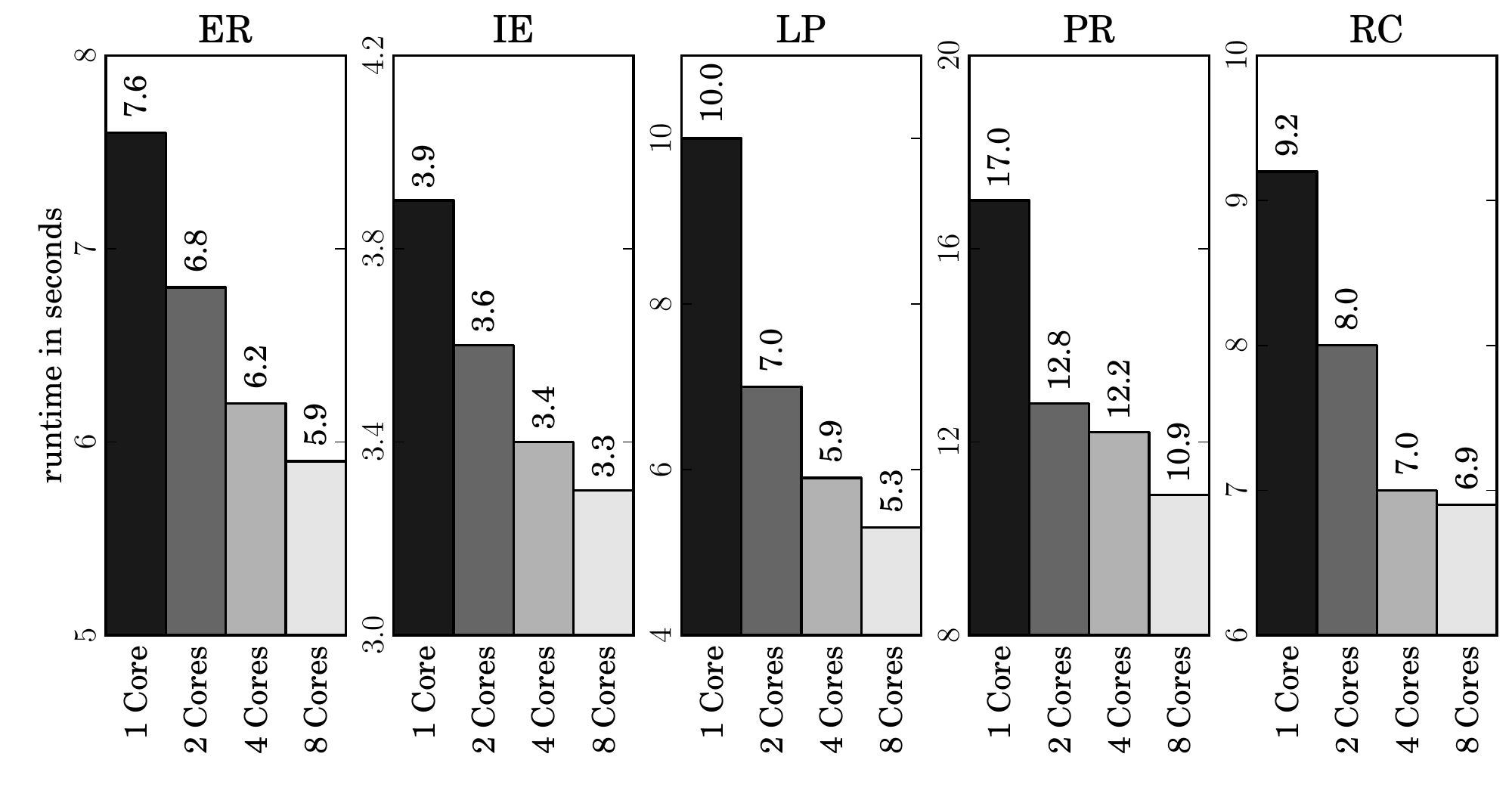}
\end{center}
\caption{\label{fig:cores}Running time of \textsc{RockIt} with CPA on varying number of cores. Due to space considerations we only show the results for the lowest gap $10^{-10}$ for ER, IE, RC, and PR benchmarks and the gap $0.01$ for the LP benchmark.}
\end{figure}

We used several established benchmark MLNs for the empirical evaluation. The entity resolution (ER) MLN addresses the problem of finding records corresponding to the same real-world entity~\cite{singla2006entity}. The information extraction (IE)~\cite{poon2007joint} MLN was created for the extraction of database records from text or semi-structured sources. The link prediction MLN (LP) was built to predict the relations holding between faculty, staff, and students of several university departments~\cite{richardson2006markov}. The protein interaction (PR) MLN was designed to predict interactions between proteins, enzymes, and phenotypes.  The relational classification (RC) MLN performs a classification on the CORA~\cite{mccallum2000automating} dataset.  In~\cite{niu2011tuffy} the MLN was used to compare the performance of \textsc{Tuffy} to the \textsc{Alchemy} system.
The ER, LP, IE, and RC MLNs were downloaded from the \textsc{Tuffy} website and the PR MLN from the \textsc{Alchemy} website.
The formula weights were learned with \textsc{Alchemy}.
Table~\ref{tab:propConstr} summarizes the properties of the five MLN benchmarks. 

Only applying CPI \textit{without} CPA already lead to significantly more compact ILPs. For the IE benchmark, for instance, the number of constraints  was reduced from $342,279$ to $4,041$. When both CPI \emph{and} CPA were used, the number of ILP constraints was further reduced for each of the MLNs. The largest reduction in the number of constraints was achieved for the PR MLN. Here, the number of constraints was reduced from $2,688,122$ to $1,573$.  Table~\ref{tab:propConstr} lists the number of ground clauses and the number of ILP constraints with CPI and with/without CPA. 

In order to compare the performance of the MLN systems, we measured the time needed to compute an interpretation whose weight (the sum in equation (1)), has a relative error of \emph{at most} $\epsilon = 10^{-\tau}, \tau \in \{1,2,3,10\},$ with respect  to the optimal weight. To this end, we used \textsc{RockIt} to compute an ILP solution  whose objective value has a relative error of at most $10^{-10}$ and computed the actual weight $\omega_{10}$ of the interpretation corresponding to this ILP solution. From this value we computed $\omega_{\tau}$ for $\tau \in \{1, 2, 3\}$ by multiplying  $\omega_{10}$ with $1-10^{-\tau}$. 
The MLN systems were run, for each $\tau \in \{1,2,3,10\}$, with an increasing number of MaxWalkSAT flips for \textsc{Alchemy} and \textsc{Tuffy} or with decreasing values of Gurobi's \textsc{MIPGap} parameter for \textsc{Markov TheBeast} and \textsc{RockIt} until a parameter configuration achieved an interpretation weight of at least $\omega_{\tau}$, or until one hour had passed, whichever came first. Figure \ref{fig:runtimesrockIt} summarizes the results for the four different gaps.

Using CPA was always more efficient except for the IE benchmark where the average running time remained almost identical. The largest decrease in running time due to CPA, from $84$ to $13$ seconds, was observed for the PR dataset. \textsc{RockIt} was more efficient and  was often able to compute a higher objective than \textsc{Tuffy}, \textsc{Markov TheBeast}, and \textsc{Alchemy}. In all cases, \textsc{RockIt} was able to compute an interpretation with highest weight in less time. We conjecture that \textsc{RockIt} without CPA is more efficient than \textsc{Markov TheBeast} because of \textsc{RockIt}'s more compact ILP formulation and the parallelization of the CPI phase.
For the ER, PR, and RC dataset \textsc{Tuffy} and \textsc{Alchemy} were not able to achieve the same approximation as \textsc{RockIt}. \textsc{Alchemy} did not finish grounding within one hour on the RC dataset and ran out of memory on the PR dataset. For the LP dataset, no system was able to achieve a gap of $0.001$ or lower.

Figure \ref{fig:cores} compares the runtime of \textsc{RockIt} with CPA for different number of cores. For each benchmark, the runtime decreases when the number of cores increases with a diminishing reduction in runtime. The LP benchmark has the highest relative decrease of about $53\%$ when comparing the running times on $1$ and $8$ cores. 

\section{Conclusions}\label{sec:conclusion}

We presented \textsc{RockIt}, a system for parallel MAP inference in SRL models combining CPI and cutting plane aggregation (CPA). CPA is a novel algorithm that aggregates symmetric constraints.
Extensive experiments showed that \textsc{RockIt} is more efficient than existing MLN systems. 
Future work will investigate more aggregation strategies and the combination of CPA with  lifted inference approaches. 

\newpage

\bibliographystyle{aaai}
\bibliography{aaai2013}

\begin{thebibliography}{}

\bibitem[\protect\citeauthoryear{Apsel and Brafman}{2012}]{apsel:2012}
Apsel, U., and Brafman, R.
\newblock 2012.
\newblock Exploiting uniform assignments in first-order mpe.
\newblock In {\em Proceedings of UAI},  74--83.

\bibitem[\protect\citeauthoryear{Asano}{2006}]{asano2006improved}
Asano, T.
\newblock 2006.
\newblock An improved analysis of goemans and williamson's lp-relaxation for
  max sat.
\newblock {\em Theoretical Computer Science} 354(3):339--353.

\bibitem[\protect\citeauthoryear{B{\"o}di, Herr, and
  Joswig}{2013}]{bodi2013algorithms}
B{\"o}di, R.; Herr, K.; and Joswig, M.
\newblock 2013.
\newblock Algorithms for highly symmetric linear and integer programs.
\newblock {\em Mathematical Programming} 137(1-2):65--90.

\bibitem[\protect\citeauthoryear{Bui, Huynh, and Riedel}{2012}]{bui:2012}
Bui, H.~H.; Huynh, T.~N.; and Riedel, S.
\newblock 2012.
\newblock Automorphism groups of graphical models and lifted variational
  inference.
\newblock {\em CoRR} abs/1207.4814.

\bibitem[\protect\citeauthoryear{Darga, Sakallah, and
  Markov}{2008}]{darga2008faster}
Darga, P.; Sakallah, K.; and Markov, I.
\newblock 2008.
\newblock Faster symmetry discovery using sparsity of symmetries.
\newblock In {\em Proceedings of the 45th annual Design Automation Conference},
   149--154.

\bibitem[\protect\citeauthoryear{de Salvo~Braz, Amir, and
  Roth}{2006}]{braz:2006}
de~Salvo~Braz, R.; Amir, E.; and Roth, D.
\newblock 2006.
\newblock {MPE} and partial inversion in lifted probabilistic variable
  elimination.
\newblock In {\em Proceedings of AAAI},  1123--1130.

\bibitem[\protect\citeauthoryear{Diaconis}{1977}]{diaconis:1977}
Diaconis, P.
\newblock 1977.
\newblock Finite forms of de finetti's theorem on exchangeability.
\newblock {\em Synthese} 36(2):271--81.

\bibitem[\protect\citeauthoryear{Domingos \bgroup et al\mbox.\egroup
  }{2012}]{Alchemy}
Domingos, P.; Jain, D.; Kok, S.; Lowd, D.; Poon, H.; and Richardson, M.
\newblock 2012.
\newblock Alchemy website.
\newblock http://alchemy.cs.washington.edu/.
\newblock last visit: 22.11.2012.

\bibitem[\protect\citeauthoryear{Finetti}{1972}]{finetti:1972}
Finetti, B.
\newblock 1972.
\newblock {\em Probability, induction and statistics: the art of guessing}.
\newblock Probability and mathematical statistics. Wiley.

\bibitem[\protect\citeauthoryear{Genesereth and
  Nilsson}{1987}]{genesereth1987logical}
Genesereth, M., and Nilsson, N.
\newblock 1987.
\newblock {\em Logical foundations of artificial intelligence}, volume~9.
\newblock Morgan Kaufmann.

\bibitem[\protect\citeauthoryear{Gogate and
  Domingos}{2011}]{gogate2011probabilistic}
Gogate, V., and Domingos, P.
\newblock 2011.
\newblock Probabilistic theorem proving.
\newblock In {\em Proceedings of UAI},  256--265.

\bibitem[\protect\citeauthoryear{Huynh and Mooney}{2009}]{huynh:2009}
Huynh, T.~N., and Mooney, R.~J.
\newblock 2009.
\newblock Max-margin weight learning for markov logic networks.
\newblock In {\em Proceedings of EMCL PKDD},  564--579.

\bibitem[\protect\citeauthoryear{Kautz, Selman, and
  Jiang}{1997}]{kautz1997general}
Kautz, H.; Selman, B.; and Jiang, Y.
\newblock 1997.
\newblock A general stochastic approach to solving problems with hard and soft
  constraints.
\newblock {\em Satisfiability Problem: Theory and Applications} 17.

\bibitem[\protect\citeauthoryear{Kersting, Ahmadi, and
  Natarajan}{2009}]{kersting:2009}
Kersting, K.; Ahmadi, B.; and Natarajan, S.
\newblock 2009.
\newblock Counting belief propagation.
\newblock In {\em Proceedings of UAI},  277--284.

\bibitem[\protect\citeauthoryear{Kisynski and Poole}{2009}]{kisynski2009lifted}
Kisynski, J., and Poole, D.
\newblock 2009.
\newblock Lifted aggregation in directed first-order probabilistic models.
\newblock In {\em Proceedings of IJCAI},  1922--1929.

\bibitem[\protect\citeauthoryear{Koch \bgroup et al\mbox.\egroup
  }{2011}]{koch2011miplib}
Koch, T.; Achterberg, T.; Andersen, E.; Bastert, O.; Berthold, T.; Bixby, R.;
  Danna, E.; Gamrath, G.; Gleixner, A.; Heinz, S.; et~al.
\newblock 2011.
\newblock Miplib 2010.
\newblock {\em Mathematical Programming Computation}  1--61.

\bibitem[\protect\citeauthoryear{Kok and Domingos}{2008}]{kok2008extracting}
Kok, S., and Domingos, P.
\newblock 2008.
\newblock Extracting semantic networks from text via relational clustering.
\newblock {\em Machine Learning and Knowledge Discovery in Databases}
  624--639.

\bibitem[\protect\citeauthoryear{Low \bgroup et al\mbox.\egroup
  }{2010}]{low:2010}
Low, Y.; Gonzalez, J.; Kyrola, A.; Bickson, D.; Guestrin, C.; and Hellerstein,
  J.~M.
\newblock 2010.
\newblock Graphlab: A new framework for parallel machine learning.
\newblock In {\em Proceedings of UAI},  340--349.

\bibitem[\protect\citeauthoryear{Lowd and Domingos}{2007}]{lowd:2007}
Lowd, D., and Domingos, P.
\newblock 2007.
\newblock Efficient weight learning for markov logic networks.
\newblock In {\em Proceedings of ECML PKDD},  200--211.

\bibitem[\protect\citeauthoryear{Margot}{2003}]{Margot:2003}
Margot, F.
\newblock 2003.
\newblock Exploiting orbits in symmetric ilp.
\newblock {\em Math. Program.} 98(1-3):3--21.

\bibitem[\protect\citeauthoryear{Margot}{2010}]{Margot:2010}
Margot, F.
\newblock 2010.
\newblock Symmetry in integer linear programming.
\newblock In {\em 50 Years of Integer Programming 1958-2008}. Springer Berlin
  Heidelberg.
\newblock  647--686.

\bibitem[\protect\citeauthoryear{McCallum \bgroup et al\mbox.\egroup
  }{2000}]{mccallum2000automating}
McCallum, A.; Nigam, K.; Rennie, J.; and Seymore, K.
\newblock 2000.
\newblock Automating the construction of internet portals with machine
  learning.
\newblock {\em Information Retrieval} 3(2):127--163.

\bibitem[\protect\citeauthoryear{Milch \bgroup et al\mbox.\egroup
  }{2008}]{milch:2008}
Milch, B.; Zettlemoyer, L.~S.; Kersting, K.; Haimes, M.; and Kaelbling, L.~P.
\newblock 2008.
\newblock Lifted probabilistic inference with counting formulas.
\newblock In {\em Proceedings of AAAI},  1062--1068.

\bibitem[\protect\citeauthoryear{Mladenov, Ahmadi, and
  Kersting}{2012}]{mladenov:2012}
Mladenov, M.; Ahmadi, B.; and Kersting, K.
\newblock 2012.
\newblock Lifted linear programming.
\newblock {\em Journal of Machine Learning Research} 22:788--797.

\bibitem[\protect\citeauthoryear{Niepert, Meilicke, and
  Stuckenschmidt}{2010}]{niepert2010probabilistic}
Niepert, M.; Meilicke, C.; and Stuckenschmidt, H.
\newblock 2010.
\newblock A probabilistic-logical framework for ontology matching.
\newblock In {\em Proceedings of AAAI},  1413--1418.

\bibitem[\protect\citeauthoryear{Niepert}{2010}]{niepert:2010}
Niepert, M.
\newblock 2010.
\newblock A delayed column generation strategy for exact k-bounded map
  inference in markov logic networks.
\newblock In {\em Proceedings of UAI},  384--391.

\bibitem[\protect\citeauthoryear{Niepert}{2012}]{niepert:2012}
Niepert, M.
\newblock 2012.
\newblock Markov chains on orbits of permutation groups.
\newblock In {\em Proceedings of UAI},  624--633.

\bibitem[\protect\citeauthoryear{Niu \bgroup et al\mbox.\egroup
  }{2011}]{niu2011tuffy}
Niu, F.; R{\'e}, C.; Doan, A.; and Shavlik, J.
\newblock 2011.
\newblock Tuffy: Scaling up statistical inference in markov logic networks
  using an rdbms.
\newblock {\em Proceedings of the VLDB Endowment} 4(6):373--384.

\bibitem[\protect\citeauthoryear{Ostrowski \bgroup et al\mbox.\egroup
  }{2011}]{ostrowski:2011}
Ostrowski, J.; Linderoth, J.; Rossi, F.; and Smriglio, S.
\newblock 2011.
\newblock Orbital branching.
\newblock {\em Math. Program.} 126(1):147--178.

\bibitem[\protect\citeauthoryear{Poole}{2003}]{poole:2003}
Poole, D.
\newblock 2003.
\newblock First-order probabilistic inference.
\newblock In {\em Proceedings of IJCAI},  985--991.

\bibitem[\protect\citeauthoryear{Poon and Domingos}{2007}]{poon2007joint}
Poon, H., and Domingos, P.
\newblock 2007.
\newblock Joint inference in information extraction.
\newblock In {\em Proceedings of AAAI},  913--918.

\bibitem[\protect\citeauthoryear{Richardson and
  Domingos}{2006}]{richardson2006markov}
Richardson, M., and Domingos, P.
\newblock 2006.
\newblock Markov logic networks.
\newblock {\em Machine learning} 62(1):107--136.

\bibitem[\protect\citeauthoryear{Riedel}{2008}]{riedel2008improving}
Riedel, S.
\newblock 2008.
\newblock Improving the accuracy and efficiency of map inference for markov
  logic.
\newblock {\em Proceedings of UAI}  468--475.

\bibitem[\protect\citeauthoryear{Shavlik and Natarajan}{2009}]{shavlik:2009}
Shavlik, J.~W., and Natarajan, S.
\newblock 2009.
\newblock Speeding up inference in markov logic networks by preprocessing to
  reduce the size of the resulting grounded network.
\newblock In {\em Proceedings of IJCAI},  1951--1956.

\bibitem[\protect\citeauthoryear{Singla and Domingos}{2006}]{singla2006entity}
Singla, P., and Domingos, P.
\newblock 2006.
\newblock Entity resolution with markov logic.
\newblock In {\em Proceedings of ICDM},  572--582.

\bibitem[\protect\citeauthoryear{Singla and Domingos}{2008}]{Singla:2008}
Singla, P., and Domingos, P.
\newblock 2008.
\newblock Lifted first-order belief propagation.
\newblock In {\em Proceedings of AAAI},  1094--1099.

\bibitem[\protect\citeauthoryear{Van~den Broeck}{2011}]{broeck:2011}
Van~den Broeck, G.
\newblock 2011.
\newblock On the completeness of first-order knowledge compilation for lifted
  probabilistic inference.
\newblock In {\em Proceedings of NIPS},  1386--1394.

\bibitem[\protect\citeauthoryear{Venugopal and Gogate}{2012}]{venugopal:2012}
Venugopal, D., and Gogate, V.
\newblock 2012.
\newblock On lifting the gibbs sampling algorithm.
\newblock In {\em Proceedings of NIPS}.
\newblock  1664--1672.

\bibitem[\protect\citeauthoryear{Wu and Weld}{2008}]{wu2008automatically}
Wu, F., and Weld, D.
\newblock 2008.
\newblock Automatically refining the wikipedia infobox ontology.
\newblock In {\em Proceedings of WWW},  635--644.

\end{thebibliography}

\end{document}